\documentclass[conference]{IEEEtran}
\IEEEoverridecommandlockouts

\usepackage{cite}
\usepackage{amsmath,amssymb,amsfonts}

\usepackage{graphicx}
\usepackage{textcomp}
\usepackage{hyperref}
\usepackage{academicons}
\usepackage{xcolor}

\usepackage{epsfig}
\usepackage{subcaption}
\usepackage{calc}
\usepackage{amssymb}
\usepackage{amstext}

\usepackage{multicol}
\usepackage{pslatex}

\usepackage{algorithm}
\usepackage{algpseudocode}

\usepackage{multirow}
\usepackage{nicefrac}

\usepackage{pgf}
\usepackage{hyperref}
\usepackage{cleveref}
\usepackage{amsfonts}
\usepackage{textcomp}
\usepackage{xcolor}

\usepackage{float}
\usepackage{longtable}
\usepackage{wrapfig}
\usepackage{tikz}
\usepackage{standalone}
\usepackage{bm}
\usepackage{bbm}
\usepackage{booktabs}
\usepackage{graphicx}

\graphicspath{ {./figs/} }

\definecolor{LightCyan}{rgb}{0.88,1,1}
\definecolor{chalkblue}{rgb}{0.671, 0.871, 0.902}
\definecolor{chalkpurple}{rgb}{0.796, 0.667,0.796}
\definecolor{chalkyellow}{rgb}{1.0, 1.0, 0.71}
\definecolor{chalkorange}{rgb}{1.0, 0.8, 0.714}
\definecolor{chalkpink}{rgb}{0.953, 0.69,0.765}

\definecolor{bohored}{RGB}{209, 133, 119}
\definecolor{bohogreen}{RGB}{130, 157, 136}
\definecolor{bohoblue}{RGB}{160, 174, 189}
\definecolor{bohoyellow}{RGB}{225, 180, 105}

\definecolor{mellowpurple}{RGB}{213, 104, 171}
\definecolor{mellowpink}{RGB}{250, 200, 220}
\definecolor{mellowgreen}{RGB}{90, 146, 91 }
\definecolor{mellowblue}{RGB}{177, 222, 220}
\definecolor{mellowyellow}{RGB}{ 254, 192, 93}
\definecolor{melloworange}{RGB}{226, 117, 76}

\definecolor{sunsetpink}{RGB}{162, 59, 85}

\definecolor{applegreen}{rgb}{0.55, 0.71, 0.0}
\definecolor{airforceblue}{rgb}{0.36, 0.54, 0.66}
\definecolor{amethyst}{rgb}{0.6, 0.4, 0.8}
\definecolor{antiquefuchsia}{rgb}{0.57, 0.36, 0.51}
\definecolor{aquamarine}{rgb}{0.5, 1.0, 0.83}
\definecolor{asparagus}{rgb}{0.53, 0.66, 0.42}
\definecolor{babyblue}{rgb}{0.54, 0.81, 0.94}
\definecolor{babyblueeyes}{rgb}{0.63, 0.79, 0.95}
\definecolor{babypink}{rgb}{0.96, 0.76, 0.76}
\definecolor{darkseagreen}{rgb}{0.56, 0.74, 0.56}
\definecolor{flavescent}{rgb}{0.97, 0.91, 0.56}
\definecolor{grannysmithapple}{rgb}{0.66, 0.89, 0.63}
\definecolor{pastelorange}{rgb}{1.0, 0.7, 0.28}
\definecolor{pastelmagenta}{rgb}{0.96, 0.6, 0.76}
\definecolor{richelectricblue}{rgb}{0.03, 0.57, 0.82}
\definecolor{rosevale}{rgb}{0.67, 0.31, 0.32}
\definecolor{sandstorm}{rgb}{0.93, 0.84, 0.25}

\definecolor{veryperi}{rgb}{10,99,255}
\definecolor{orchidbloom}{rgb}{194,166,245}

\definecolor{popcorn}{cmyk}{3,13,53,0}
\definecolor{bubblegum}{cmyk}{3,67,25,0}
\definecolor{daffodil}{cmyk}{0,29,76,0}
\definecolor{poinciana}{cmyk}{14,89,92,4}
\definecolor{harborblue}{cmyk}{87,37,44,27}
\definecolor{Cascade}{cmyk}{57,4,36,0}
\definecolor{spunsugar}{cmyk}{33,1,7,0}
\definecolor{coccamocha}{cmyk}{36,48,55,34}
\definecolor{fragilesprout}{cmyk}{34,13,93,1}
\definecolor{supersonic}{cmyk}{95,64,10,1}

\newcommand{\R}{\mathbb{R}}
\newcommand{\Z}{\mathbb{Z}}
\newcommand{\C}{\mathbb{C}}

\newcommand{\F}{\mathcal{F}}

\newcommand{\cin}{\text{in}}
\newcommand{\out}{\text{out}}

\renewcommand{\i}{\mathrm{i}}
\renewcommand{\O}{\mathcal{O}}

\newcommand\AB[1]{\textcolor{black}{#1}}

\newcommand\orcid[2]{
\hspace*{-1.5mm}$^\text{, #2}$\thanks{$^\text{#2}$https://orcid.org/#1}\hspace*{-1mm}
}

\def\BibTeX{{\rm B\kern-.05em{\sc i\kern-.025em b}\kern-.08em
    T\kern-.1667em\lower.7ex\hbox{E}\kern-.125emX}}
    
\begin{document}

\title{LFA applied to CNNs: Efficient Singular Value Decomposition of Convolutional Mappings by Local Fourier Analysis
\thanks{This work is partially supported by the German Federal Ministry for Economic Affairs and Climate Action, within the project “KI Delta Learning” (grant no.\ 19A19013Q). M.R.\ acknowledges support by the German Federal Ministry of Education and Research within the junior research group project “UnrEAL” (grant no.\ 01IS22069). The contribution of K.K.\ is partially funded by the European Union’s HORIZON MSCA Doctoral Networks programme project AQTIVATE (grant no.\ 101072344).}
}

\author{
\IEEEauthorblockN{Antonia van Betteray$^{\text{1}}$\orcid{0000-0002-2338-1753}{a},
Matthias Rottmann$^{\text{1}}$\orcid{0000-0003-3840-0184}{b} and
Karsten Kahl$^{\text{1}}$\orcid{0000-0003-3840-0184}{c}}
\IEEEauthorblockA{
\textit{$^{1}$IZMD, University of Wuppertal, Germany} \\
\{vanbetteray, rottmann, kkahl\}@uni-wuppertal.de\\} }

\maketitle

\begin{abstract}
The singular values of convolutional mappings encode interesting spectral properties, which can be used, e.g., to improve generalization and robustness of convolutional neural networks
as well as to facilitate model compression. However, the computation of singular values is typically very resource-intensive. The naive approach involves unrolling the convolutional mapping along the input and channel dimensions into a large and sparse two-dimensional matrix, making the exact calculation of all singular values infeasible due to hardware limitations. In particular, this is true for matrices that represent convolutional mappings with large inputs and a high number of channels. Existing efficient methods leverage the Fast Fourier transformation (FFT) to transform convolutional mappings into the frequency domain, enabling the computation of singular values for matrices representing convolutions with larger input and channel dimensions. For a constant number of channels in a given convolution, an FFT can compute $N$ singular values in $\mathcal{O}(N \log N)$ complexity. In this work, we propose an approach of complexity $\mathcal{O}(N)$ based on local Fourier analysis, which additionally exploits the shift invariance of convolutional operators. We provide a theoretical analysis of our algorithm's runtime and validate its efficiency through numerical experiments. Our results demonstrate that our proposed method is scalable and offers a practical solution to calculate the entire set of singular values -- along with the corresponding singular vectors if needed -- for high-dimensional convolutional mappings.

\end{abstract}


\section{{Introduction}}
\label{sec:introduction}

Deep convolutional neural networks (CNNs) are powerful methods for image recognition tasks~\cite{russakovsky_imagenet_2015,krizhevsky_imagenet_2012,he_deep_2016}.  Their core operations are convolutional mappings~\cite{lecun_gradient-based_1998}, which perform a linear transformation of their inputs to extract features. The spectral properties of convolutional mappings have diverse applications. The spectral norm is used for regularization in order to improve generalizability~\cite{yoshida_2017,sedghi2018singular,gouk2021} or improve robustness to adversarial attacks \cite{Cisse2017,chen_backdoors_2024}. Furthermore the SVD can be used for low-rank approximation to enable model compression \cite{Jaderberg2014SpeedingUC,zhang_accelarating2015,ben-noach-goldberg-2020-compressing,chen_groupreduce_2018,acharya_online_2019,hsu2022languagemodelcompressionweighted}, or to study interpretability in CNNs~\cite{Praggastis2022TheSO}. 
Pseudo-invertible neural networks are designed to learn both, a task and its inverse~\cite{hsu2022languagemodelcompressionweighted}. Notably, the pseudo-inverse can be computed directly via singular value decomposition (SVD).

The weight tensor corresponding to a convolutional mapping is a 4D tensor, comprising dimensions of the input and output channels and the filter height and width. During the convolution, each 2D filter slides across the input feature map. At each position, the operation computes the sum of element-wise multiplications between the filter and the corresponding overlapping region of the input, producing output feature maps.

\begin{figure}
\centering
  \begin{subfigure}[t]{0.45\linewidth}
  \centering
       \includegraphics[width=.9\textwidth]{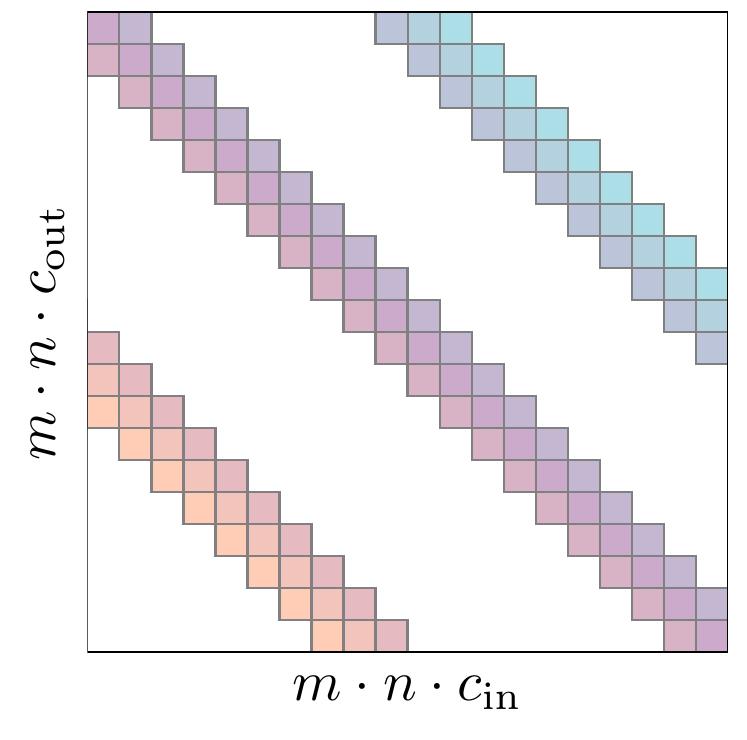}
       \caption{Matrix view of $3 \times 3$ convolution on $m \times n$ input.}
       \label{fig:conv_matrix}
  \end{subfigure}\hfill
  \begin{subfigure}[t]{0.45\linewidth}
  \centering
      \includegraphics[width=.9\textwidth]{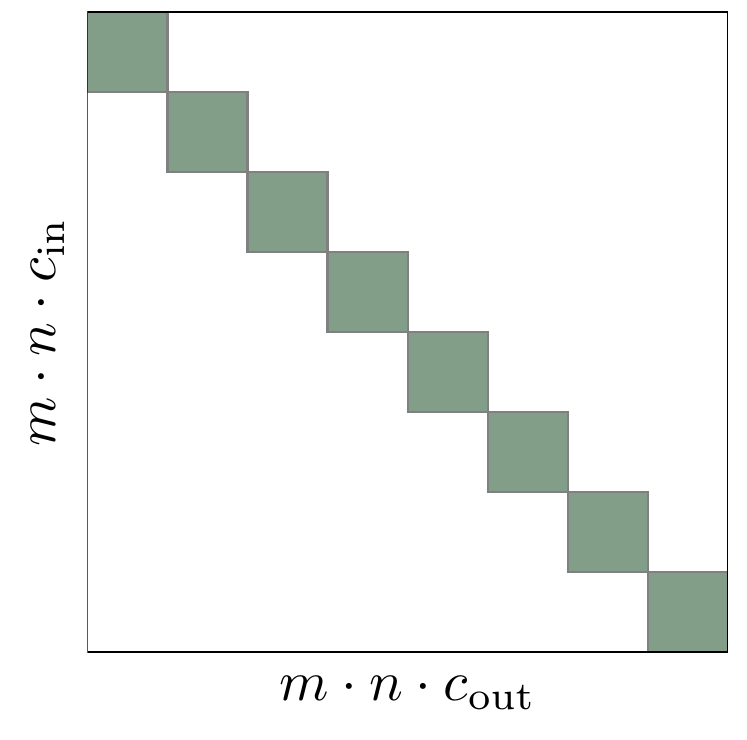}
      \caption{Matrix composed of $m \cdot n$ blocks of size $c_{\cin} \times c_\out$.}\label{fig:block_matrix}
  \end{subfigure}
\end{figure}

Considering an input with spatial dimensions $m \times n$ and {$c_{\cin}$, $c_{\out}$ denoting the number of input and output channels, respectively,} the convolution mapping is given by

\begin{equation}
    A : \R^{m \times n \times c_{\cin}} \rightarrow \R^{m \times n \times c_{\out}} \, .
\end{equation}

The corresponding 2D matrix is sparse with dimension $(m \cdot n \cdot c_{\cin}) \times (m \cdot n \cdot c_{\out})$ with sparsity pattern according to~\cref{fig:conv_matrix}.
It is obvious, that the size of this matrix grows rapidly as the number of channels and/or the input size increases, whereas applying the SVD on large matrices is time and memory consuming. 
Assuming a square input size and equal number of channels, i.e., $c:=c_{\cin}=c_{\out}$ and $n=m$, the computational complexity for the brute-force approach is $\O(n^6 c^3)$ \cite{sedghi2018singular}. 
The focus of this work is on an efficient approach to reduce the computational complexity of computing SVD of large matrices corresponding to convolutional weight tensors. 
The view of a convolution as a doubly circulant matrix allows for the use of (discrete) Fourier transform. By applying the 2D Fast Fourier Transform (FFT) the computational complexity can be reduced to $\mathcal{O}(n^2 c^2(c + \log n))$~\cite{sedghi2018singular}. On the other hand our approach takes into account, that convolutional mappings are translation invariant, which allows for an application of local Fourier Analysis~\cite{kahl_automated_2020}. Local Fourier Analysis is primarily used in the study of iterative methods for solving partial differential equations (PDEs), especially multigrid (MG) methods. Structural analogies of CNNs and MG has been analyzed by~\cite{he_deep_2016,he_mgnet_2019,eliasof_mgic_2020,van_Betteray_2023_ICCV,van_Betteray_2025}. 
We utilize the convolution theorem, which states that convolutions diagonalize under Fourier transforms. 
In our approach we specify a complex Fourier exponential as an ansatz function for the basis change into the frequency domain. We take advantage of this by transforming the analysis into the frequency domain -- via Fourier transforms -- where the action of these matrices on different frequency components (i.e., Fourier modes) can be analyzed independently.
Specifically, the orthogonality of the Fourier basis functions ensures that a convolutional operator in the spatial domain is converted into a pointwise multiplication in the frequency domain under the Fourier transform. Consequently, the convolutional mapping based on a 4D weight tensor is transformed into a 2D block-diagonal matrix, cf.\ \cref{fig:block_matrix}.
Each block corresponds to a specific frequency component and has dimensions $c_{\out}\times c_{\cin}$, coupling only across channels. 
The operator remains diagonal along the spatial dimensions $n$ and $m$.
This block structure allows for independently calculating SVD for smaller matrices, which reduces the computational complexity to $\O(n^2 c^3)$, compared to the FFT-based approach which requires $\O(n^2 c^2 ( c + \log n))$.

To preserve spatial dimension of the input after the application of the convolution and effectively process the edge pixels of the input image, the  input typically is padded by zeros around the border. In the context of PDEs, zero padding is referred to Dirichlet boundary conditions. While zero padding is the standard for image recognition tasks, the application of local Fourier Analysis as well as FFT assumes periodic boundary conditions for the convolution. That raises the question of how the boundary conditions influence the accuracy of the spectrum of the convolutional mapping.

The contribution of this paper can be summarized as follows.
\begin{itemize}
    \item We propose an algorithm for efficient computation of singular values of convolutional mappings by exploiting translation invariance of convolutional operators that allows for the application of local Fourier Analysis. 
    \item We provide a theoretical analysis of our algorithm's runtime and validate the efficiency through numerical experiments.
    \item We study the influence of Dirichlet and periodic boundary conditions to analyze and compare the similarities of the resulting spectra.
\end{itemize}
{Our method can improve the efficiency and scalability whenever a given task requires the computation of a singular value decomposition of convolutional mappings.}

The remainder of this paper is structured as follows. \Cref{sec:related_works}
reviews related works, including methods for determining spectra or approximating the largest singular values, as well as selected applications.
In \cref{sec:svd_with_lfa} we introduce our method by first presenting lattices and crystalline structures and their relevance to convolutional mappings. This foundation leads to the presentation of the convolutional theorem, from which we derive our algorithm and analyze its computational complexity. \Cref{sec:results} validates and studies the proposed approach in terms of numerical results and their discussion. Finally, \cref{sec:conclusion} concludes the paper by summarizing our contributions and findings.

\section{{Related Works}}\label{sec:related_works}
In this section, we review related works that analyze spectral properties of convolutions. These works can be broadly categorized into 1) methods of exact computation of the full spectrum, 2) approximation of the (largest) singular values 3) applications. For the sake of readability, throughout the present section let $m=n$ and $c := c_{\cin} = c_{\out}$. Furthermore, we assume square-shaped kernels of extent $k \times k$.

\paragraph{Exact Computation of the Full SVD} 
An approach to calculate the complete spectrum is introduced by~\cite{sedghi2018singular}. They characterize the singular values of a convolutional layer by using $c^2$ FFTs followed by $n^2$ SVDs. The FFTs require $\O (n^2 \log n)$ flops while the SVD uses $\O(c^3)$, resulting in an overall computational complexity of $\O(n^2 c^2(c + \log n))$. This approach is closely related to ours, which is based on the calculation of the {so-called} symbol, requiring $n^2 \cdot \O(1)$. Compared to \cite{sedghi2018singular}, our approach reduces {the} computational complexity {of} the calculation of the singular values by a factor $\O(\log n)$ to {$\O(n^2 c^3)$}.

\paragraph{Approximation of the Largest Singular Value} 
Yoshida and Miyato~\cite{yoshida_2017} reshape the weight tensor into a dense $c \times c \cdot k^2$ matrix and use a power iteration method to approximate the spectral norm, i.e.\ the largest singular value. However, the reshaped matrix yields only $c \cdot k^2$ singular values, whereas the corresponding linear transformation of the convolution has $c \cdot n^2$ singular values.  Nevertheless, they utilize the operator norm as a regularizer, demonstrating that the approximation obtained by their representation is sufficient for the given purpose, though being a loose upper bound, see also~\cite{Cisse2017,Tsuzuku2018LipschitzMarginTS}.
An efficient method to compute the $L^1$ and $L^\infty$ norms of convolutional layers was proposed by~\cite{gouk2021} to approximate their spectral norm. In case of convergence, their method computes the exact spectral norm of a convolutional layer. However, does this not provide insights into the whole spectrum of the layer.

\begin{figure*}[t]
\centering
    \begin{subfigure}[t]{.45\textwidth}
        \includegraphics[width=\textwidth]{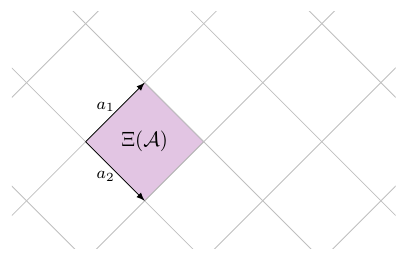}
        \caption{Unit cell $\Xi(\mathcal{A})$ corresponding $\mathbb{L}(\mathcal{A})$.}\label{fig:unitcell}
    \end{subfigure}
    \begin{subfigure}[t]{.45\textwidth}
        \includegraphics[width=\textwidth]{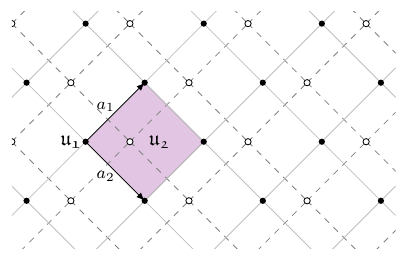}
        \caption{Crystal $\mathbb{L}^{\mathfrak{u}}(\mathcal{A})$.}\label{fig:crystal}
    \end{subfigure}
    \caption{\Cref{fig:unitcell} Lattice $\mathbb{L}(\mathcal{A})$ with basis $\mathcal{A} = [a_1\, a_2]$.\cref{fig:crystal} A crystal $\mathbb{L}^\mathfrak{u}(\mathcal{A})$ contains copies of $\mathcal{A}$, each shifted by $\mathfrak{u_1}, \ldots, \mathfrak{u}_\nu$.}\label{fig:lattice_crystal}
\end{figure*}

\begin{figure*}[t]
\centering
    \begin{subfigure}[t]{.45\textwidth}
        \includegraphics[width=\textwidth]{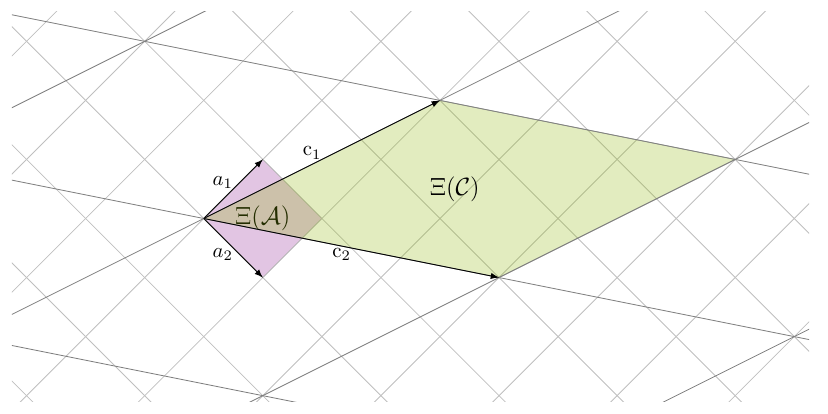}
        \caption{Sublattice with primitive vectors $c_1$ and $c_2$.}\label{fig:sublattice}
    \end{subfigure}
    \begin{subfigure}[t]{.45\textwidth}
        \includegraphics[width=\textwidth]{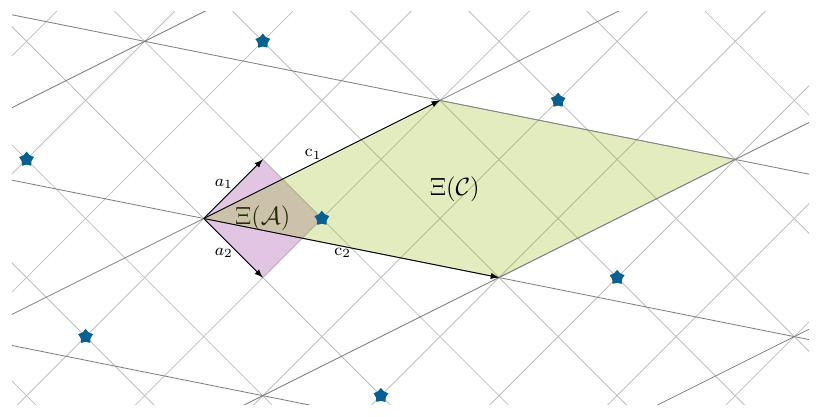}
        \caption{Crystal $\mathbb{L}^{\mathfrak{u}}(\mathcal{A})$.}\label{fig:crystaltorus}
    \end{subfigure}
    \caption{
{Sublattice $\mathbb{L}(\mathcal{C}) \subseteq \mathbb{L}(\mathcal{A})$ with $\mathcal{C} =\mathcal{A} \cdot \begin{bmatrix} 
    3 & 2 \\
    1 & 3     
\end{bmatrix}$ 
and the periodicity of the corresponding crystal torus $\mathbb{T}_{\mathcal{A},\mathcal{C}}$.}\label{fig:sublattice_latticetorus}
}
\end{figure*}

\paragraph{Application Domains}
The ability to compute the SVD of convolutions paves the way to multiple applications. For instance, singular values can be utilized for the regularization of CNNs, enhancing the generalizability. Regularizers based on the spectral norm are studied by~\cite{sedghi2018singular,yoshida_2017,gouk2021,neyshabur2018a} and more recently by \cite{singla2021fantastic,pmlr-v97-ryu19a}.

The spectral norm was also utilized by~\cite{Cisse2017,chen_backdoors_2024} to improve the robustness to adversarial attacks.
Instead of focusing only on the largest singular value, other domains require access to a broader portion of the spectrum.
 
The SVD can be used for low rank approximations to enable model compression. Extensive studies on low rank approximation in neural networks have been conducted by~\cite{Jaderberg2014SpeedingUC,denton_exploiting2014,zhang_accelarating2015,ben-noach-goldberg-2020-compressing}. These methods have been further developed to compress word embedding layers \cite{chen_groupreduce_2018,acharya_online_2019,hsu2022languagemodelcompressionweighted}.
The application of the SVD in CNN interpretability is studied by~\cite{Praggastis2022TheSO}.
Also, in~\cite{van_Betteray_2025} the minimal and maximal singular values are utilized for the initialization of polynomial CNN blocks.

Recently,~\cite{bolluyt_psinn} introduced a class of pseudo-invertible neural networks designed to learn both a task and its inverse. They leverage the inverse operation as a method for image generation, formulating the pseudo-inverse
$B = A(AA^T)^{-1}$ of a convolution $A$ as a transposed convolution. Rather than computing the exact inverse, they restructure CNN layers in order to approximate it. Alternatively, an exact inverse can be achieved through efficient SVD computation.

\section{SVD with Local Fourier Analysis}\label{sec:svd_with_lfa}
In this section we establish our approach to efficiently calculate the SVD using LFA. 
In order to better prepare the reader to the following discussion we start by a short introduction to lattices and crystals, following~\cite{kahl_automated_2020}.
An (ideal) crystal is characterized by the infinite repetition of a basic structural unit, the unit cell, arranged in a regular pattern defined by a lattice, cf.\ \cref{fig:lattice_crystal}. Although {our experiments in} this work focus on rectangular grids defined by images, crystalline structures can come in other shapes, i.e.\ octagonal, which allows for a generalization of our method.
The introduction to lattices and crystals is followed by a review of convolutional mappings, the presentation to the convolution theorem and our proposal to its application within LFA. Eventually we summarize our approach in an algorithm, whose computational complexity is examined by analyzing its time complexity. 

\paragraph{Introduction to Lattices and Crystals}

Let $\mathcal{A} : = [a_1 \; a_2] \in \R^{2 \times 2}$ be a set of {linearly independent {vectors, so-called} \textit{primitive vectors}, $a_1, a_2$}, also known as lattice basis. The set of points

\[
\mathbb{L} = \{ x = \mu_1 a_1 + \mu_2 a_2 \in \R^2 \} \text{ with } \mu_1, \mu_2 \in \Z
\]

is a $2$-dimensional lattice $\mathbb{L}$ generated by $\mathcal{A}$. In matrix 
notation $\mathbb{L}$ is defined as 

\begin{equation*}
    \mathbb{L} (\mathcal{A} ) := \mathcal{A}\mathbb{Z}^2 = \{ x = \mathcal{A} \cdot \mu : \mu \in \mathbb{Z}^2\}.
\end{equation*}

A corresponding \textit{unit cell} is given by 

\[
\Xi(\mathcal{A}) = \mathcal{A}[0, 1)^2= \{ x = \mathcal{A} \cdot \varphi : \varphi \in [0,1)^2\}.
\]

\Cref{fig:lattice_crystal}\,(a)
depicts a $2$-dimensional lattice with lattice basis $a_1$ and $a_2$ and the respective unit cell \AB{$\Xi(\mathcal{A})$}.
The union over all lattice elements $ x \in \mathbb{L}(\mathcal{A})$ satisfies

\[
\dot \cup_{x \in \mathbb{L}(\mathcal{A})}\{ x + \xi : \xi \in \Xi(\mathcal{A)} \}= \R^2.
\] 

Specifically, the union of all unit cells covers all of $\R^{2}$ {without overlap} and each unit cell can be associated with one lattice point $x \in \mathbb{L}(\mathcal{A})$. A generalization of the concept of lattice structures is given by \textit{crystal} structures, which allow us to work with multiple points $\mathfrak{u}_{1},\ldots,\mathfrak{u}_{\nu} \in \Xi(\mathcal{A})$, within each unit cell, 

\[
\mathbb{L}^{\mathfrak{u}}(\mathcal{A} )  \{x = \mathcal{A}\cdot \mu + \mathfrak{u}_{\ell}\, :\, \mu \in \mathbb{Z}^{2} \text{\ and\ } \ell = 1,\ldots, \nu\}.
\] 

\Cref{fig:crystaltorus} illustrates the structure elements $\mathfrak{u}_1$ and $\mathfrak{u}_2$ associated with the unit cell $\Xi(\mathcal{A})$.

That is, in contrast to lattices, crystals can describe arbitrary arrangements of points within a unit cell, which are then translated according to $\mathbb{L}(\mathcal{A})$. Equivalently, one could also say that a crystal structure is a collection of lattices $\mathbb{L}(\mathcal{A})$ which are translated w.r.t.\ one another by $\mathfrak{u}_{1},\ldots,\mathfrak{u}_{\nu}.$ Finally, in order to restrict a lattice or crystal structure to a finite extent we introduce the concept of lattice and crystal tori, which are defined by the quotient group

\[
\mathbb{T}_{\mathcal{A},\mathcal{C}} = \mathbb{L}^{\mathfrak{u}}(\mathcal{A})\Big\slash \mathbb{L}(\mathcal{C}),
\] 

where $\mathbb{L}(\mathcal{C}) \subseteq \mathbb{L}(\mathcal{A})$ is any sublattice of $\mathbb{L}(\mathcal{A})$. Thus there exists an integer matrix $Z \in \mathbb{Z}^{2\times 2}$ with $\mathcal{C} = Z\cdot \mathcal{A}$. According to~\cite{kahl_automated_2020} the number of degrees of freedom in $\mathbb{T}_{\mathcal{A},\mathcal{C}}$ is then given by $|\operatorname{det}(Z)|\cdot |\mathfrak{u}|$ with $|\mathfrak{u}|$ denoting the number of degrees of freedom in each unit cell of the crystal $\mathbb{L}^{\mathfrak{u}}(\mathcal{A}).$ 

\Cref{fig:sublattice} visualizes an example sublattice $\mathbb{L}(\mathcal{C})$ and \cref{fig:crystaltorus} illustrates the corresponding lattice torus $\mathbb{T}_{\mathcal{A, C}}$.

\paragraph{Convolutional Mappings in CNNs}

The situation found in CNNs for image processing can be mapped to the notation of lattices and crystals as follows. On each layer of the network we expect an input with spatial extent of size $n\times m$ with $c_{\cin}$ channels, fixing the distance between two orthogonally adjacent spatial entries to one, this input can be thought of as living on the crystal structure

\[
\mathbb{T}^{\mathfrak{c}_{\cin}}_{n,m} = \mathbb{L}^{\mathfrak{c}_{\cin}}\big(\begin{bmatrix}1 & 0\\0 & 1\end{bmatrix}\big)\Big\slash \mathbb{L}\big(\begin{bmatrix}m & 0\\0 & n\end{bmatrix}\big),
\] 

where $\mathfrak{c}_{\cin}$ accounts for the fact that at each spatial location in the input we find $c_{\cin}$ collocated channel degrees of freedom. Note, that we distinguish between $\mathfrak{c_{\cin}}$ denoting the position of these channel degrees of freedom and $c_{\cin} = | \mathfrak{c}_{\cin}|$ their number. Likewise the output of a convolutional layer can be thought of living on the structure

\[
\mathbb{T}^{\mathfrak{c}_{\out}}_{n,m} = \mathbb{L}^{\mathfrak{c}_{\out}}\big(\begin{bmatrix}1 & 0\\0 & 1\end{bmatrix}\big)\Big\slash \mathbb{L}\big(\begin{bmatrix}m & 0\\0 & n\end{bmatrix}\big).
\] 

Note, that taking this point of view we imply the use of periodic boundary conditions, which albeit not necessarily natural, simplifies the upcoming discussion of Fourier analysis of convolutions. It is easy to see that the number of degrees of freedom in the respective crystal tori corresponds exactly to the input and output sizes. Within this framework a convolutional mapping

\begin{equation}
    A: \R^{m \times n \times c_{\cin}} \rightarrow \R^{m \times n \times c_{\out}}
\end{equation} 

can also be thought of as a mapping

\[
A: \mathcal{L}(\mathbb{T}^{\mathfrak{c}_{\cin}}_{n,m}) \longrightarrow \mathcal{L}(\mathbb{T}^{\mathfrak{c}_{\out}}_{n,m}).
\] 

\begin{figure}[h]
\centering
\includegraphics[width=.35\linewidth]{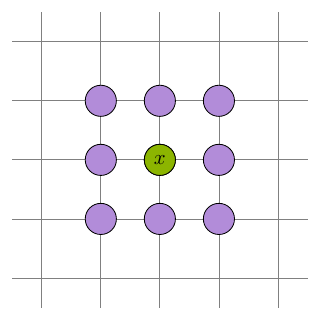}
\caption{$3\times3$ kernel operator $\mathcal{N}$, centered at $x$ (green dot).}\label{fig:stencilcoordinates}
\end{figure}

To be more precise the convolution can then be written as a multiplication operator, which acts on an input $f$ in the following way

\begin{align*}
    (A \ast f)(x)  & = \sum_{y\,\in\, \mathcal{N}} M_{y}\cdot f(x+y).
\end{align*} In here $\mathcal{N}$ describes the extent of the kernel operator $A$ and is typically assumed to be local, e.g., the $3\times 3$ neighborhood centered at $x$ as shown in \cref{fig:stencilcoordinates}. The multiplication matrices $M_{y}$ are of size $ c_{\out}\times c_{\cin}$.

\begin{table*}[ht]
  \caption{
  Time complexity of \cref{alg:svd_with_lfa} for computing the SVD with our proposed LFA-based approach, in comparison to the time complexity {of the} FFT~\cite{sedghi2018singular} as well as {of} the SVD of the explicit matrix representation of the convolution mapping.}
    \label{tab:timecomplexity_of_svd_with_lfa}
    \centering
    \scalebox{.9}{
    \begin{tabular}{c | c|c | c}
    \toprule
    Method & spatial input dimension & channel dimension & time complexity \\ 
    \midrule
    \midrule
    explicit & $m=n$& $c = c_{\cin} = c_{\out}$ & $\O(n^6 c^3)$ \\[1ex]
        FFT & $m=n$ & $c = c_{\cin} = c_{\out}$ & $\O(n^2 c^2 (c + \log n))$ \\
    \midrule
    LFA (ours)& $n = m$  & $c = c_{\cin} = c_{\out}$ &  $\O(n^2 c^3)$ \\[1ex]
     & $n \neq m$  & $c = c_{\cin} = c_{\out}$ &  $\O(n m c^3)$ \\[1ex]
     & & $  c_{\cin} \geq c_{\out}$ &  $\O(n m c_{\cin}^2 c_{\out})$ \\[1ex]
     & & $  c_{\cin} \leq c_{\out}$ &  $\O(n m c_{\cin} c_{\out}^2)$ \\
  \bottomrule
    \end{tabular}}
\end{table*}

\paragraph{Convolution Theorem}

As pointed out in~\cite{kahl_automated_2020}, the introduced crystal structures allow us to introduce wave functions $f_{k} = \operatorname{e}^{2\pi \i \langle k,x \rangle }$ with $x \in \mathbb{T}_{n,m}$, where the frequencies $k$ that lead to well defined functions are given by

\[
    \mathbb{T}^{\star}_{n,m} = \mathbb{L}\big(\begin{bmatrix}\tfrac{1}{n} & 0\\0 & \tfrac{1}{m}\end{bmatrix}\big)\Big\slash \mathbb{L}\big(\begin{bmatrix}1 & 0\\0 & 1\end{bmatrix}\big)
\] 

The extension to the crystal case is straight-forward and a canonical basis of wave functions can be fixed. Applying a convolution $A$ to these wave functions immediately yields

\[
(A\ast f_{k})(x) = \big(\sum_{y\,\in\, \mathcal{N}} M_{y}\cdot \operatorname{e}^{2\pi \i \langle k,y \rangle }\big) \operatorname{e}^{2\pi \i \langle k,x \rangle }.
\]

That is, the wave functions a.k.a.~Fourier modes yield invariant subspaces w.r.t.~the convolution $A$. Note, that due to the fact that the convolution acts on a crystal structure with $c_{\cin}$ elements per unit cell in the pre-image space and $c_{\out}$ elements per unit cell in the image space there is a basis of $c_{\cin}$ or $c_{\out}$ Fourier basis functions for each frequency $k\in \mathbb{T}^{\star}_{n,m}$. For brevity we omit the explicit description here and refer to the derivation in~\cite{kahl_automated_2020} for more details. 

Collecting an orthonormal basis of all Fourier modes for a particular frequency as columns of matrices $\mathbf{F}^{\mathfrak{c}_{\cin}}_{k}$ and $\mathbf{F}^{\mathfrak{c}_{\out}}_{k,c_{\out}}$, respectively, the above statement can be rephrased interchangeably to

\[
A_k \mathbf{F}^{\mathfrak{c}_{\cin}}_{k} = \mathbf{F}^{\mathfrak{c}_{\out}}_{k} A_{k}
\] 

where the symbol $A_{k}$ of $A$ at frequency $k$ is given by the $c_{\out}\times c_{\cin}$ matrix

\[
A_{k} = \sum_{y\,\in\, \mathcal{N}} M_{y}\cdot \operatorname{e}^{2\pi \i \langle k,y \rangle }.
\] 

Due to the fact that $\mathbf{F}^{\mathfrak{c}_{\cin}}_{k}$ and $\mathbf{F}^{\mathfrak{c}_{\out}}_{k}$ have orthonormal columns, we thus can calculate a singular value decomposition of $A$ by first computing the decomposition

\[
A_{k} = U_{k} \Sigma_{k} V_{k}^{\star}
\]
for each $k \in \mathbb{T}^{\star}_{n,m}$. Then 

\[
\widehat{U}_{k} = \mathbf{F}^{\mathfrak{c}_{\out}}_{k} U_{k} \text{\ and\ } \widehat{V}_{k} = \mathbf{F}^{\mathfrak{c}_{\cin}}_{k} V_{k}
\] have orthonormal columns and are left and right singular vectors of $A$ with corresponding singular values contained in $\Sigma_{k}$. Collecting all $\widehat{U}_{k}, \Sigma_{k}$ and $\widehat{V}_{k}$ for all $k\in \mathbb{T}^{\star}_{n,m}$ then yields the full SVD. Most importantly the set of all singular values of the convolution can be computed without forming the global singular vectors $\widehat{U}_{k}$ and $\widehat{V}_{k},$. Computing the singular values in this fashion alleviates the dependency of this computation on the spatial dimensions $n$ and $m$, thus drastically reducing the computational complexity. This process is summarized in~\cref{alg:svd_with_lfa}.

\begin{algorithm}
\caption{SVD with LFA($A, m, n$)}\label{alg:svd_with_lfa}
\begin{algorithmic}[1]
\State \textbf{Init} {\; $X = \{0, \frac{1}{n}, \frac{2}{n}\ldots, \frac{n-1}{n}\}$,
$Y =  \{0, \frac{1}{m}, \frac{2}{m}\ldots, \frac{m-1}{m}\}$,
 $K = X \times Y $\;}
 \For{$i\; = 1, \ldots, n$}
   \For{$j = 1, \ldots, m$} 
       \State $k = K_{i, j}$ 
       \State $B_{i, j} = 
       \sum_{y \in \mathcal{N}} M_y \cdot e^{2 \pi \texttt{i} \langle k, y \rangle}$
       \State $U_{i,j}, \, \Sigma_{i, j},\, V^\ast_{i,j} = \operatorname{SVD}(B_{i, j})$\;
   \EndFor
     \EndFor
\end{algorithmic}
\end{algorithm}

\paragraph{Computational Complexity} 
The time complexity of our SVD calculation with LFA, cf.\ \cref{alg:svd_with_lfa} will be analyzed in the following. 
\Cref{alg:svd_with_lfa} is comprised of two $\operatorname{\mathbf{for}}$-loops with
time complexity of $\O(n)$ and $\O(m)$ respectively, cf.\ line 1 and 2, followed by a Fourier transform in line $4$ with time complexity of $\O(1)$. The time complexity of the SVD in line $5$ depends on the channel dimensions, since $B_{i,j} \in \C^{c_{\cin} \times c_{\out}}$. For an equal number of input and output channels $c$ the time complexity of SVD for dense matrices is $\O(c^3)$~\cite{trefethen_numerical_2022}.
Summarized, the overall computational complexity is $\O(n^2c^3)$ if $n =m$. 
Time complexity for unequal numbers of input and output channels are specified in~\cref{tab:timecomplexity_of_svd_with_lfa}.

\begin{figure}[ht]
\begin{minipage}{.45\linewidth}
\includegraphics[width=\textwidth]{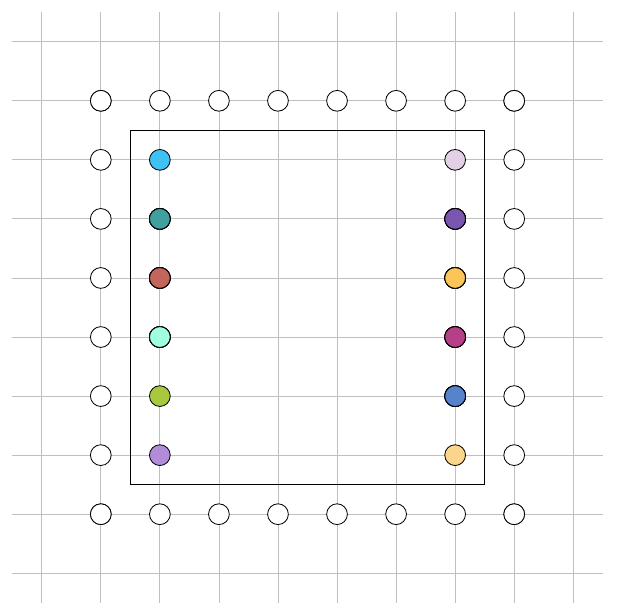}
\end{minipage}
\hfill
\begin{minipage}{.45\linewidth}
    \includegraphics[width=\textwidth]{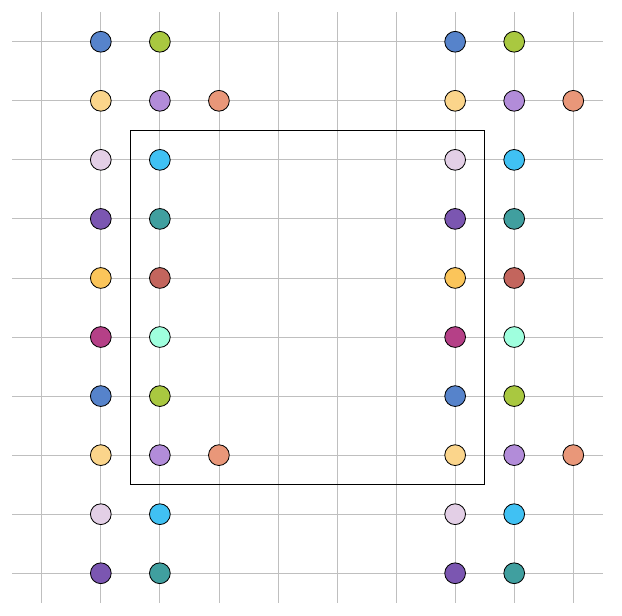}
\end{minipage}
 \caption{Dirichlet boundary conditions (left), the white dots symbolize zeros vs.~periodic boundary conditions (right).}\label{fig:periodic_bounds}
\end{figure}

\paragraph{Boundary Conditions}
{The LFA that is used to compute the singular values of the convolution operators implicitly requires periodic boundary conditions, exemplarily illustrated in~\cref{fig:periodic_bounds}. Though in many applications, these convolutions are formulated with Dirichlet boundary conditions, i.e., zero padding. In~\Cref{sec:results} we take a closer look at the distribution of singular values for both types of boundary conditions to gauge their impact.}

\section{Numerical Results}\label{sec:results}
\begin{figure*}[h]
    \centering
    \scalebox{0.55}{
    \input{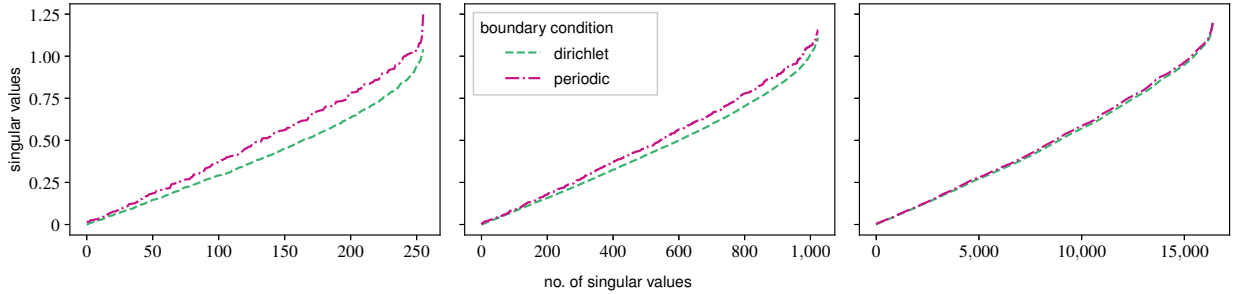}  
    }
    \caption{Effect of boundary conditions for increasing input sizes ($n=4, 8, 32$; left to right). The number of input and output channels is fixed to $16$.}
    \label{fig:boundary_condition}
\end{figure*}

In this section we present and discuss numerical results.
We start by analyzing the dependence of our approach on the given boundary conditions in order to study the applicability of our approach to CNNs, that are typically implemented with zero padding{, i.e. Dirichlet boundary conditions}.
The computational complexity of our algorithm was already studied in~\cref{sec:svd_with_lfa}. 
In this section, our theoretical analysis is complemented with experiments on the computational efficiency of our method in terms of 
runtime to calculate the SVD. If not mentioned otherwise, we focus on the case of square matrices. Note that, the corresponding (unrolled) matrices have full rank, so that the number of singular values equals the number of columns and rows, respectively. This section concludes with a discussion on how the memory layout effects the computational runtime.

\paragraph{Boundary Conditions}
The application of LFA requires periodic boundary conditions. However, typically convolutions are padded with zeros which in the PDE-perspective corresponds to Dirichlet boundary conditions. In the following we ignore the requirement of periodic padding and study the influence on the boundary conditions on the spectrum. To this end, we compare the result of our LFA-based method, which computes singular values under the assumption of periodic boundary conditions with the results of the naive baseline, i.e., explicitly setting up the sparse matrix that corresponds to the convolution with zero padding, from which the singular values are computed using NumPy~\cite{2020NumPy-Array}.
\Cref{fig:boundary_condition} {depicts} the singular values for $3$ weight tensors, each with $16$ input and output channels {over a range of input sizes} $n$. Clearly, with increasing input size the number of singular values increases proportionally. We observe that for a small number of singular values / small input size $n$, the boundary conditions clearly affect the approximation quality of our method. However, this effect disappears with increasing number of singular values to be computed. This effect is also to be expected, as the boundary has less influence for growing lattice sizes. Consequently, for larger lattice sizes our method yields useful approximations to the singular values of convolutions with zero padding.

\paragraph{Runtime Analysis}
To demonstrate the {computational efficiency} of our algorithm, we compare the execution runtime (in seconds $s$) for calculating singular values using our algorithm against both the brute-force approach and FFT-based method proposed by~\cite{sedghi2018singular}.
We use an Intel(R) Xeon(R) Gold 6242 CPU $@ 2.80 GHz$ and all $16$ cores are available during the calculation. We start the timing before the transformation of the weight tensor with FFT and LFA respectively and stop after the calculation of the singular values. Our implementation operates on PyTorch convolutional weight tensors, which are stored in a channel-first format. To ensure correct alignment of the spatial dimensions, the weight tensors must therefore be transposed prior to applying the FFT. The computational cost of this transposition, in terms of runtime, is independent of the input dimension and  is negligible at less than $8\times 10^{-6}$ seconds. For both approaches, we utilize the $\operatorname{svd}$ function from NumPy's $\operatorname{linalg}$ module with option $\operatorname{compute\_ uv=False}$.

We begin our discussion by analyzing the execution time required to calculate all singular values by the FFT-based method and our LFA-based approach, and compare it to the naive explicit approach, which computes the singular values of the sparse matrix obtained when explicitly representing the convolutional operator as matrix.

The largest matrix we decomposed by explicitly unfolding the convolutional mapping was of $65,\!536 \times 65,\!536$, corresponding to a convolutional operator with $16$ in- and output channels convolved with an input with dimension $n= 64$. Beyond that, memory capacity becomes quickly a limiting factor. \Cref{fig:runtimes_lfa_fft_dense} shows that the runtime for the explicit approach grows rapidly for increasing values of $n$. For example, the computation of the singular values of a $1,\!024 \times 1,\!024$ matrix takes $0.30$ seconds, while it takes over $400$ seconds to compute the singular values for a matrix of size $16,\!384 \times 16,\!384$. For very small values of $n$, i.e.\ $n=4$ and $n=8$, the FFT-based approach is the fastest. However, as $n$ increases, its runtime grows more quickly than that of our LFA-based approach. Consequently, for $n  \geq 16$ the computation of the singular values is faster with our LFA-based approach than with the FFT. With $n=256$, we observe that the FFT-based approach requires $2.51$ seconds, while our LFA-approach requires $2.30$, resulting in a speed-up factor of $1.09$, cf.~\cref{tab:runtime_ratio}.

\begin{figure*}
    \begin{subfigure}[t]{.485\textwidth}
    \scalebox{.68}{
        \input{figs/pgfs/runtime-lfa_vs_fft-loglog.pgf}}
       \caption{Comparison of execution runtimes for computing singular values using a naive approach with the explicit matrix representation, an FFT-based approach and our LFA method. The evaluated matrices are square, with both channel dimensions fixed to $16$, while the input feature map size varies as $n \in \{4, 8,16, 32, 64, 256, 512\}$.   Both axes are on a logarithmic scale.}\label{fig:runtimes_lfa_fft_dense}
       \end{subfigure}
       \hfill
        \begin{subfigure}[t]{.485\textwidth}
         \scalebox{.68}{%
         \input{figs/pgfs/runtime-lfa_vs_fft-xlog.pgf}     
         }
    \caption{Execution runtime for computing the SVD using our LFA approach compared to the FFT-based approach. 
    Channel dimensions are fixed to $16$ and the input feature map size varies with $n \in \{2^{i}: i = 8, \ldots, 14\} $. The $x$-axis shows the corresponding total number of singular values in millions (M) on a logarithmic scale.
    }\label{fig:runtime_lfafft_best}
    \end{subfigure}
\end{figure*}

This trend continues as the input size $n$ further increases, which is also shown in~\cref{fig:runtime_lfafft_best}. Here we compare the runtime of our LFA-based approach to the runtime of the FFT-based approach for $n \geq 256$. Especially for very large $n$, {the gap in runtime between our LFA-based approach and the FFT becomes more pronounced}. In the case of $n=16,\!384$, the SVD computation involves over $4$ million (M) singular values. Their computation requires approximately $181$ minutes with the FFT-based approach, whereas our LFA-based approach requires only about $125$ minutes, i.e.\ we achieve a speed-up factor of $1.44$, cf.\ \cref{tab:runtime_ratio}. Overall, as to be expected from our theoretical runtime analysis, the speed-up factor increases as $n$ increases.

\begin{table}[t]
\caption{Ratio $\nicefrac{(s_{\text{FFT}}}{s_{\text{LFA}})}$ of runtime to calculate the SVD after transformation by FFT  and LFA. ($s_{\text{FFT}}$) denotes the overall runtime for the FFT-based approach and  ($s_{\text{LFA}}$) for our LFA-based approach.} 
    \centering
    \scalebox{.8}{
    \begin{tabular}{r r |c  r r }
    \toprule
$n$ &   no. of SVs &  method &  ${\text{runtime }} (s)$ & $\nicefrac{s_{\text{FFT}}}{s_{\text{LFA}}}$\\
    \midrule
     \midrule
 $256$ & $1,\!048,\!576$ & FFT & $ 2.51 $ &  \\
  & & LFA & $ 2.30 $ & $ 1.09$ \\ 
  \midrule
 $512$ & $4,\!194,\!304$ & FFT & $ 9.96 $ &  \\
  & & LFA & $ 8.49 $ & $ 1.17 $ \\ 
  \midrule
$1024$ & $16,\!777,\!216$ & FFT & $38.00 $ &  \\
  & & LFA & $ 33.01 $ & $ 1.15 $ \\ 
  \midrule
$2048$ & $67,\!108,\!864$ & FFT & $ 149.96 $ & \\
     &   & LFA & $ 130.20 $ & $ 1.15 $ \\ 
  \midrule
 $4096$ & $268,\!435,\!456$ & FFT & $596.44 $ & \\
  & & LFA & $ 520.21 $ & $ 1.15 $ \\ 
  \midrule
 $8192$ & $1,\!073,\!741,\!824$ & FFT & $ 2535.28 $ & \\
  & & LFA & $2077.44 $ & $ 1.22 $ \\ 
  \midrule
  $16384$ &  $4,\!294,\!967,\!296$ & FFT & $10,864.97 $ & \\
  & & LFA & $ 7,521.93 $ & $ 1.44 $ \\ 
 \bottomrule
\end{tabular}}
\label{tab:runtime_ratio}
\end{table}

\paragraph{Memory Layout Effects}
Our proposed method to efficiently calculate the singular values of a weight tensor convolved with some input feature map consists of two steps: 
\begin{enumerate}
    \item Transformation by LFA (or FFT resp.).
    \item Computation of the singular values.
\end{enumerate}
Both the Fourier transforms, the FFT and the LFA, yield $n^2$ block matrices of dimension $c \times c$, each treated separately by the SVD routine. Given the identical dimensions and data types of the transformed tensors, one intuitively expects the SVD computation to require the same computational effort for the overall LFA and FFT. However, in practice, we observed that the computation time required for computing the SVD differs for both methods, as shown in~\cref{tab:runtimes_transformation+svd}. We found that the memory layouts of the tensors obtained after transformation with FFT and LFA differ. In general python stores data in row-major format.\footnote{Other orders such as column-major format are also possible but are not relevant in this work.} In earlier experiments we found, that ensuring a row-major memory layout prior to the FFT yields a faster total runtime for $n < 8,\!192$, and prior to the LFA $n \leq 8,\!192$, respectively. This memory layout conversion is independent of the input dimension and is negligible for the total runtime.\footnote{The results reported in this work corresponds to the fastest runtimes achieved for both methods, LFA and FFT, regardless of memory layout.} 
Additionally the memory layout is maintained during the transformation-part of the LFA in our implementation. However, this does not hold for the transformation performed by the FFT routine of NumPy, i.e., which does not return row-major memory layout, even if the input is ensured to be row-major.
\Cref{tab:runtimes_transformation+svd} reports an overview of the runtimes for the two steps involved in efficiently computing the SVD: the transformation time ($s_{\F}$) and the computation time of SVD ($s_{\text{SVD}}$), which together sum up to the total computation time ($s_{\text{total}}$). As already pointed out, the runtimes for computing the SVD differ for the same input dimensions for both approaches. Furthermore, it shows, that the row-major memory layout in our LFA-approach does not only improve the runtime for the transformation LFA but also the improves the computation time of the subsequent SVD. To ensure that the computation of the SVD is equally fast for both approaches, we converted the tensors obtained after the FFT into row-major memory layout. For comparison, we include experiments, for our LFA-based approach, where the data is only converted to row-major after the LFA, rather than beforehand. The results are presented in~\cref{tab:runtimes_ascontigous}. Corresponding to observations made in~\cref{tab:runtimes_transformation+svd}, a row-major memory layout enables more efficient SVD computation. 
According to~\cref{tab:runtimes_ascontigous}, for $n \leq 8,\!192$, it is possible to convert the memory layout of the transformed tensors, such that after the transformation part of the FFT and the LFA, the respective subsequent SVD computations require about the same runtime. Although computing the SVD in row-major memory layout is faster than in the layout produced by the FFT, the overhead of converting to row-major layout outweighs the benefit, making it more efficient to compute the SVD directly on the layout obtained by the FFT. This effect is amplified for large $n$. Consequently, our approach not only enables a faster transformations by LFA but also preserves a memory layout which facilitates efficient SVD computations.

\begin{table}[t]
   \centering
       \caption{Runtime for computing the singular values for different values of $n$. The total runtime ($s_{\text{total}}$) consists of the transformation time ($s_{\F}$) and the time required to compute of SVD ($s_{\text{SVD}}$).}
    \label{tab:runtimes_transformation+svd}
\scalebox{.8}{
    \begin{tabular}{r r  c | c cc}
    \toprule
        $n$ & no.\ of SVs & method ($\F$) & $s_{\F}$ & $s_{\text{SVD}}$ & $s_{\text{total}}$ \\
         \midrule
         \midrule
        $16,\!384$ &  $4,\!294,\!967,\!296$ & FFT & $1,\!966.89$ & $8,\!898.08$ & $10,\!864.97$ \\[1ex]
        & & LFA & $1,\!595.02$ & $5,\!926.91$ & $7,\!521.93$ \\
        \midrule
        $8,\!192$ &  $1,\!073,\!741,\!824$ & FFT & $317.95$ & $2,\!217.33$  & $2,\!535.28$ \\[1ex]
        &  & LFA & $82.48$  & $1,\!994.97$ & $2,\!077.44$\\
         \midrule
        $4,\!096$ & $268,\!435,\!456$ & FFT & $69.11$ & $527.33$ & $596.44$ \\[1ex]
          & & LFA & $20.57$ & $499.64$ & $520.21$ \\
         \midrule
        $2,\!048$ & $67,\!108,\!864$ & FFT & $18.19$ & $131.77$ & $ 149.96 $ \\[1ex]
                  &                  & LFA & $4.97$ & $125.23$ &  $130.20 $  \\ 
         \midrule
        $1,\!024$ & $16,\!777,\!216$ & FFT & $4.86$ & $33.14$ & $ 38.00 $ \\[1ex]
                  &                  & LFA & $1.28$ & $31.74 $ &  $33.01$  \\ 
        \bottomrule
    \end{tabular}}
\end{table}

\begin{table}[t]
\caption{Effect of row-major layout on the runtimes $s_\F$, $s_{\text{SVD}}$ and $s_{\text{total}}$. If the input for the transformation $\F$ is in row-major layout it is marked with \checkmark and otherwise with $\times$.  Recall, that row-major layout is not maintained by the FFT. The time required to convert to row-major layout is reported under $s_\text{copy}$. A dash (--) indicates that no conversion was performed.}
\centering
   \scalebox{.8}{
    \begin{tabular}{r  c | c  cccc}
   \toprule
    $n$ &  $\F_{\text{method}}$ & row-major & $s_{\F}$ &  $s_{\text{copy}}$ &  $s_{\text{SVD}}$ &$s_{\text{total}}$ \\
    \midrule
    \midrule

  $8,\!192$   & FFT & $\times$  &  $317.95$&  --      & $2,\!217.33$   & $2,\!535.28$ \\   
             & FFT & $\checkmark$  & $333.18$& $558.88$& $1,\!986.25$   & $2,\!878.31$ \\
             & LFA & $\checkmark$ & $82.48$ & --      & $1,\!994.97$   & $2,\!077.44$ \\
              & LFA & $\times$ & $87.37$ &$1,\!021.24$& $1,\!987.23$   & $3,\!095.83$ \\
              \midrule
              $4,\!096$ & FFT & $\checkmark$ & $68.96$ & -- & $527.62$ &  $596.58$\\
              & FFT & $\checkmark$  & $67.44$ & $63.48$ & $497.35$    & $628.28$  \\ 
              & LFA & $\checkmark$   & $20.57$ &  --     & $499.64$    & $520.21$  \\
              & LFA & $\times$  & $22.59$ & $148.56$& $497.01$    & $668.16 $ \\
              \midrule
  $2,\!048$  & FFT &  $\checkmark$  & $17.88$ & -- & $131.73$ & $149.61 $ \\\
              & FFT &  $\checkmark$  & $18.16$ & $13.81$ & $125.61$    & $157.58$  \\         
   & LFA &  $\checkmark$  & $4.97$  & --      & $125.23$    & $130.20$ \\ 
              & LFA   & $\times$ & $5.73$  & $30.89$ & $125.47$    & $162.10$  \\ 
              \midrule
    $1,\!024$ & FFT & $\checkmark$ & $4.87$  & --      & $33.07$     & $ 37.94$ \\ 
              & FFT & $\checkmark$  & $4.87$  & $3.07$  & $31.89$     & $39.84$  \\
              & LFA & $\checkmark$  & $1.28$  & --      & $31.74 $    & $33.01$  \\
              & LFA & $\times$  & $1.43$  & $7.00$  & $31.76$     & $40.08$  \\ 
         \bottomrule
    \end{tabular}}
    \label{tab:runtimes_ascontigous}
\end{table}

\section{Conclusion}\label{sec:conclusion}

In this work we presented an efficient algorithm to compute the singular values of convolutional operators corresponding to large sparse matrices. In contrast to previous approaches, our LFA-based method utilizes the translation invariance of convolutional operators to achieve optimal scaling of the computational effort with the spatial resolution $n$ of the convolution. We provide a theoretical analysis to prove this. Compared to a previous FFT-based approaches, our LFA-based method improves the computational complexity by a factor $\log(n)$.
We undergird this theoretical improvement by runtime studies. 
We verified that the time required to compute the SVD decreases as
$n$ increases. Moreover, we found, that our LFA implementation produces a memory layout that is advantageous for the subsequent computation of the SVD, leading to a further runtime reduction. We make our code publicly available at \url{https://github.com/vanbetteray/conv_svd_by_lfa}. It is worth noting that, unlike the FFT, the LFA is embarrassingly parallel.

\bibliographystyle{ieeetr}
{\small
\bibliography{lit, lit2}
}

\end{document}